%% file: ijcai22-demo-track.tex
\documentclass{article}
\usepackage{ijcai22}

\usepackage{times}

\usepackage{soul}
\usepackage{url}
\usepackage[hidelinks]{hyperref}
\usepackage[utf8]{inputenc}
\usepackage[small]{caption}
\usepackage{graphicx}
\usepackage{amsmath}
\usepackage{booktabs}
\usepackage{xkeyval} 
\usepackage{graphicx}
\usepackage{amsmath}
\usepackage{booktabs}
\usepackage{multirow}
\usepackage{hyperref}
\usepackage{comment}
\usepackage{xspace}
\usepackage[colorinlistoftodos]{todonotes}
\usepackage{enumitem}
\usepackage{microtype}
\usepackage{float}
\usepackage{natbib}

\newcommand{\ITMo}{\texttt{ITMo}\xspace}
\newcommand{\Ha}{\texttt{HANS}\xspace}
\newcommand{\TMKG}{\texttt{TMKG}\xspace}
\newcommand{\CSKG}{\texttt{CSKG}\xspace}

\newcommand{\OSM}{OpenStreetMap\xspace}

\newcommand{\QT}[1]{``{#1}''\xspace}

\title{Intelligent Traffic Monitoring with Hybrid AI}

\author{
Ehsan Qasemi$^1$\footnote{Work performed during internship at Bosch.}\and
Alessandro Oltramari$^2$
\\
\affiliations
$^1$University of Southern California, Los Angeles, CA, USA\\
$^2$Bosch Research and Technology Center, Pittsburgh, PA, USA\\

\emails
qasemi@usc.edu,
alessandro.oltramari@us.bosch.com
}

\begin{document}

\maketitle

\input{Sections-demo/Abstract}

\section{Introduction}
\label{sec:Intro}
\input{Sections-demo/Introduction}

\section{Intelligent Traffic Monitoring: Problem Formulation}
\label{sec:problem}
\input{Sections-demo/Problem}

\section{Human-Assisted Neuro-Symbolic Architecture for Multi-modal Context Understanding (\Ha)}
\label{sec:arch}
\input{Sections-demo/Architecture}

\section{Evaluation}
\label{sec:demo}
\input{Sections-demo/Demonstrator}

\section{Conclusion}
\label{sec:conclusion}
\input{Sections-demo/Conclusions}


\bibliographystyle{named}
\bibliography{ijcai22}






\end{document}

%% file: Sections-demo/Abstract.tex
\begin{abstract}
    Challenges in Intelligent Traffic Monitoring (\ITMo) are exacerbated by the large quantity and modalities of data and the need for the utilization of state-of-the-art (SOTA) reasoners.
    We formulate the problem of \ITMo and introduce \Ha, a neuro-symbolic architecture for multi-modal context understanding, and its application to \ITMo.
    \Ha utilizes knowledge graph technology to serve as a backbone for SOTA reasoning in the traffic domain.
    Through case studies, we show how \Ha addresses the challenges associated with traffic monitoring while being able to integrate with a wide range of reasoning methods.
\end{abstract}

%% file: Sections-demo/Introduction.tex
U.S.\ Intelligent Traffic Management sector is estimated to be worth approximately \$3B\@ and will expand with a CAGR\footnote{Compound Annual Growth Rate.} forecast to be 10.2\% through 2028~\cite{GrandView21}.
This projection is likely to be adjusted to reflect the recent approval of the U.S.\ infrastructure bill, set to infuse an additional \$11B in transportation safety, including programs to reduce crashes and fatalities~\cite{CNN21}.

Intelligent Traffic Monitoring (\ITMo) represents an essential instrument to improve road safety and security in intelligent transportation systems~\cite{pascale2012wireless,9112118}.
Moreover, with the advent of autonomous vehicles, \ITMo is poised to become an even more relevant part of the smart city infrastructure of the future~(the abundance of real-time data points can be used to make traffic mitigation strategies
more efficient and effective~\cite{chowdhury2021towards}).

\ITMo focuses on deriving actionable knowledge from a large array of networks of sensors deployed along highways, city roads and intersections, etc.
\citet{muppalla2017knowledge} use knowledge graph technology for traffic congestion detection by proposing an ontology based on SSN~\cite{haller2019modular} to interpret the video feed from traffic cameras and a heuristic reasoner on top of the ontology.
Similarly, \citet{lam2017real} use real-time image processing to count the number of tail-lights in the video feed from traffic cameras.
However, both works fall short in accommodating a wide range of multi-modal and background information on traffic. 
Transformer-based language models (LMs) are SOTA methods in many natural language understanding tasks \cite{liu2019roberta} and show human-level commonsense knowledge~\cite{ma2019towards-hykas}.
However, to the best of our knowledge, no effort has been make to integrate them into \ITMo.

Our \textbf{first} contribution is formulating the problem of \ITMo in presence of large quantities of multi-modal, background (including commonsense), and domain knowledge that can leverage SOTA symbolic and neural reasoners.

Our \textbf{second} contribution is \Ha, a neuro-symbolic architecture for multi-modal context understanding in \ITMo.
We evaluate \Ha, by implementing set reasoning case studies, where each showcases the capability of \Ha to process different modalities of data or integrate with various of type of reasoners.
Our results show that \Ha, can serve as a platform for the integration of SOTA in \ITMo and paves the way for future research in the field.

%% file: Sections-demo/Problem.tex
We formulate the \ITMo as the problem of deriving actionable knowledge from multi-modal real-time and background knowledge.
We identify six main requirements for \ITMo, all related to processing and reasoning over high volumes of heterogeneous data.
\begin{enumerate}
    \item \textbf{Multi-modal information fusion}.
    \ITMo involves data processing at scale, which can include transient information generated by multiple surveillance cameras, traffic lights, speed radars, acoustic sensors, etc. ~\cite{lancelle2016distributed}. 
    As aggregating such data is inherently complicated, generalizing over them to obtain high-level traffic situations constitutes an open problem.
    
    \item \textbf{Background Knowledge Acquisition}.
    Any human-like reasoning performed by AI systems over traffic situations should include knowledge of the common features that can be encountered in most typical scenarios. 
    For instance, we know that a car coming to a stop on the side of a highway is either malfunctioning or indicating an emergency experienced by the driver;
    we also know that the presence of a police vehicle parked right behind the car changes the scene interpretation, suggesting that the driver was prompted to pull over due to an alleged traffic violation.
    Commonsense knowledge stems from everyday experience~\cite{sap2019atomic,ilievski2020consolidating}: it can be causal, e.g.\ vehicles typically stop in front of the red light; spatial, e.g.\ two cars cannot occupy the exact same location, or even cultural, e.g.\ the so-called  \textit{Pittsburgh left}\footnote{A colloquial term for the driving practice of the first left-turning vehicle taking precedence over vehicles going straight through an
    intersection, associated with the Pittsburgh, Pennsylvania, area.}.
    
    \item \textbf{Domain Knowledge Extension}.
    Public knowledge can also be leveraged for \ITMo, e.g., the map of a geographical region of interest, e.g. \OSM~\cite{OpenStreetMap}, live traffic observed by drivers, e.g. using the Waze app\footnote{\hyperlink{https://www.waze.com}{https://www.waze.com}}, relevant weather information, e.g.\ OpenWeatherMap\footnote{\hyperlink{https://openweathermap.org/}{https://openweathermap.org/}}, or historical traffic data, e.g.\ SigAlert~\cite{walton2011livelayer}. 
    Combining all these information sources requires an advanced semantic model, capable of federating knowledge at different levels of granularity.   
    \item \textbf{Flexibility}.
    Any reasoning system for \ITMo must have a flexible design to allow for incremental development, which requires testing the algorithms across different traffic scenarios, factoring in additional modalities, etc.
    \item \textbf{Integration with SOTA}.
    \ITMo must allow for seamless integration with SOTA machine learning (e.g.,
    computer vision algorithms) and inference engines, both at the symbolic and neural level;
    \item \textbf{Robustness}.
    \ITMo must be able to handle noisy information, such as wrong semantic labels from
    computer vision algorithms, missing video frames, etc.
\end{enumerate}

To satisfy these six requirements, we propose a neuro-symbolic architecture that integrates knowledge graph technologies with neural networks.

%% file: Sections-demo/Architecture.tex
\begin{figure}
    \centering
    \includegraphics[width=\columnwidth]{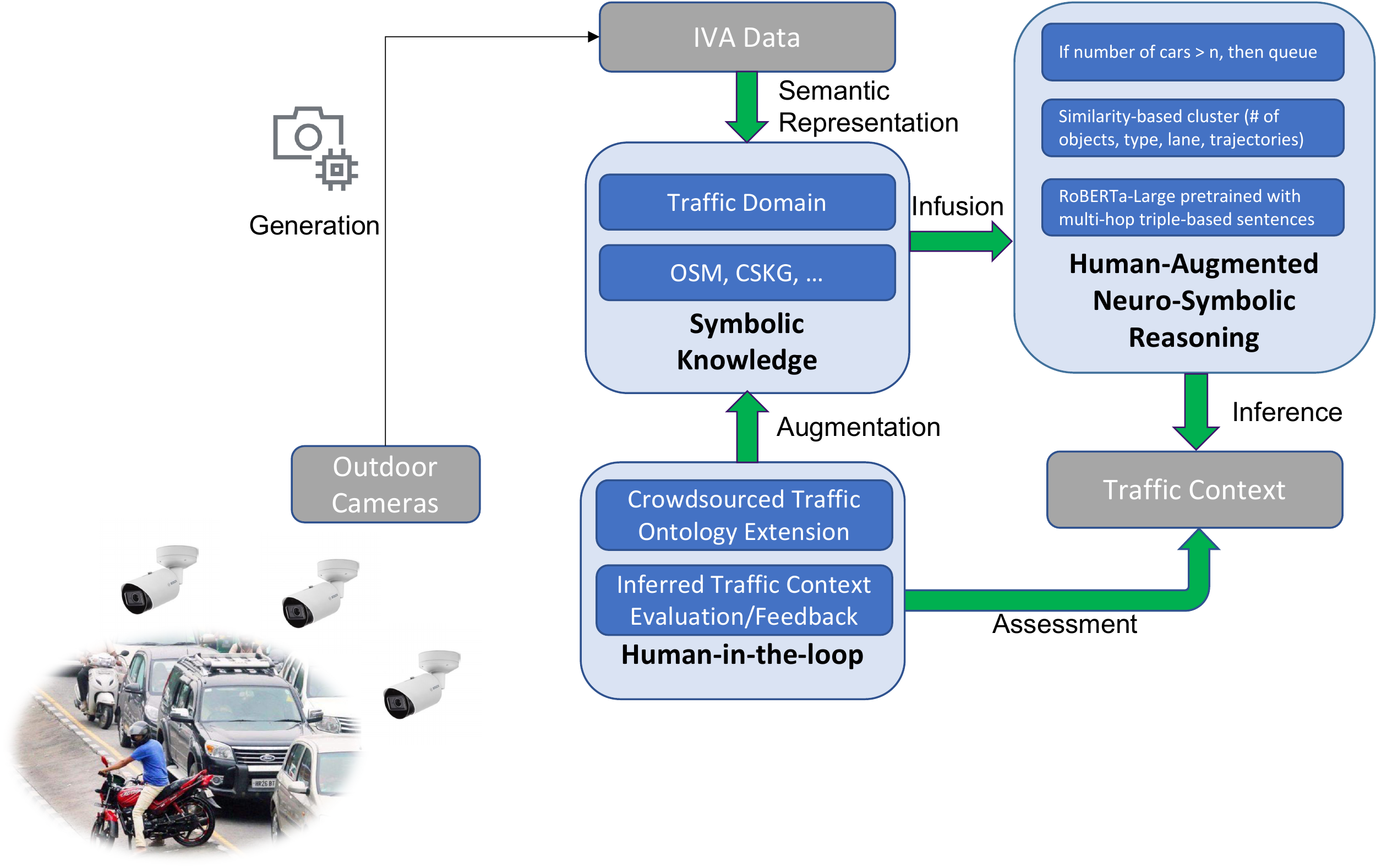}
    \caption{Human-Assisted Neuro-Symbolic Architecture for Multi-modal Context Understanding}
    \label{fig:hans-a}
    
\end{figure}

\Ha has been designed to easily adapt to various sensor-based domains, with the ultimate goal of enhancing road security and safety.
In the following, we briefly describe the architecture pipeline shown in figure~\ref{fig:hans-a}.

\paragraph{Generation.}
To extract information from images of the camera feed, we use SOTA off-the-shelf computer vision models. 
Here, we used object detection models to process individual frames from videos collected by stationary surveillance cameras.
This process, which is illustrated in figure~\ref{fig:hans-a} as IVA data includes: object classes~(car, truck, pedestrian, etc.)~\cite{he2015deep,detr}, speed and trajectory of moving objects~\cite{rad2010vehicle}, relative position (based on road topology and lane information)~\cite{ventura2018iterative}.

To extract the background information and domain knowledge, we relied heavily on existing resources of public knowledge, e.g. Wikidata~\cite{vrandevcic2014wikidata} and \OSM~\cite{OpenStreetMap}. 

\paragraph{Semantic Representation.}
An ontology for traffic monitoring (\textit{domain level}) is built to extend the
Scene Ontology (\textit{core level}) developed by Bosch~\cite{tiddi2020neuro} based on W3C's Semantic Sensor Network ontology specifications ~\cite{haller2019modular} (\textit{foundational level}).
Through suitable mapping mechanisms, the IVA data are used to instantiate the domain ontology, generating the Traffic Monitoring Knowledge Graph (\TMKG)\footnote{Implemented using Stardog, a well-known enterprise-level knowledge graph
framework - see \url{https://www.stardog.com/}}.
This is further expanded with relevant partitions of two different resources: the federated
Commonsense Knowledge Graph (\CSKG), to provide commonsense background knowledge related to the traffic
domain~\cite{ilievski2020consolidating}, and Open Street Maps~\cite{OpenStreetMap}, to expand traffic scenes based on the location-based knowledge consistent with the GPS coordinates of the cameras.

\paragraph{Augmentation \& Assessment.}
These modules include two main human-in-the-loop processes: an early phase where relevant concepts to model the traffic domain are elicited using crowdsourcing, and a late phase where crowd workers are asked to assess the results
of human-augmented neuro-symbolic reasoning.
\paragraph{Infusion \& Inference.}
In these two intimately connected modules, the \TMKG is used as the basis for
reasoning, in particular by
i) triggering rule-based inferences for simple predictions (e.g., a given number of cars on a lane may be considered as a
threshold to identify a~\textit{traffic queue};
ii) training similarity-based statistical algorithms with specific triple-based scene features~(type of objects,
trajectory, speed, etc.);
iii) fine-tuning a language model with sentences generated from multi-hop (edge) connections linking different \TMKG nodes.
The current demo\footnote{See video clip at this link:\url{https://tinyurl.com/476nwf2n}} is based on a sample dataset\ (2 days' worth of video footage) collected from a single stationary camera deployed over an intersection in Chesapeake, Virginia.
The demo ~(Figure~\ref{fig:hans-d}) and relevant examples are discussed in section~\ref{sec:demo}.
 
%% file: Sections-demo/Demonstrator.tex
\Ha is designed with ease of reasoner integration in mind: here we showcase this key feature by demonstrating different types of
inference methods.
First, we focus on conventional rule-based models that aim to find traffic congestion in video feeds.
Second, we present two advanced approaches, respectively centered around \TMKG and a pre-trained language model.

\begin{figure}[ht]
    \centering
    \includegraphics[width=\columnwidth]{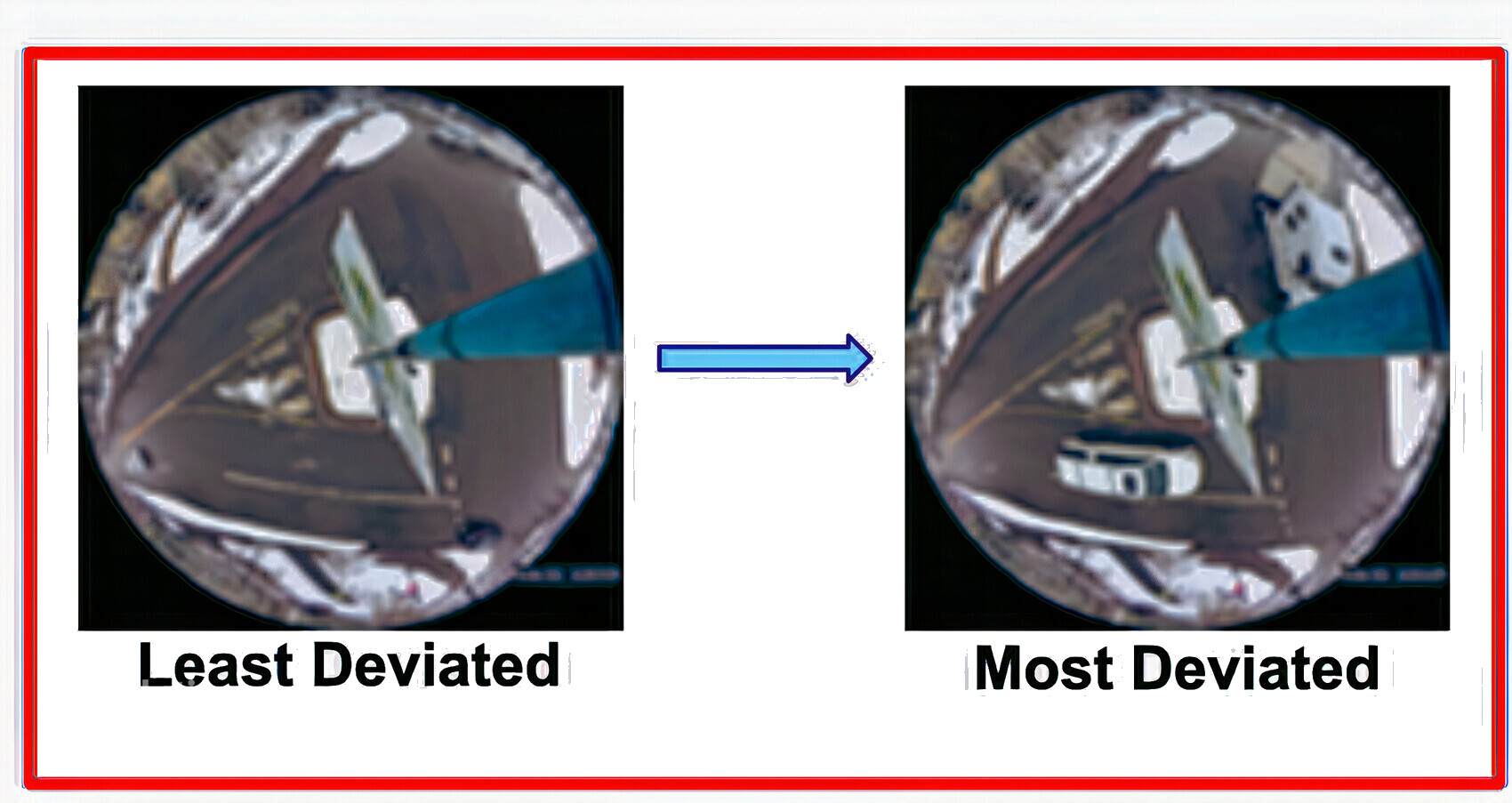}
    \caption{Baseline rule-based method for traffic congestion detection. Left: the least deviated frame from
    the median scene. Right: the most deviated frame from the median scene. The former has lower probability of congestion than the latter.}
    \label{fig:deviation}
\end{figure}

\paragraph{Symbolic Method.}
Explainability~\footnote{The degree which humans achieve deep understanding of the internal procedures that take place while the model is training or making decisions.\cite{linardatos2020explainable}} of results and human interpretability~\footnote{the ability to explain or to present in understandable terms to a human \cite{doshi2017towards}} make symbolic methods a popular choice in critical applications such as \ITMo~\cite{muppalla2017knowledge,lam2017real}.
\citet{muppalla2017knowledge} use the deviation of each frame of video from the median as a core indicator of traffic congestion.
Figure~\ref{fig:deviation} illustrates the results of \Ha's implementation of this method, in which we have two frames from the video feed with, respectively, least and most deviation from the median frame.

\begin{figure*}[ht]
    \centering
    \includegraphics[width=0.52\linewidth]{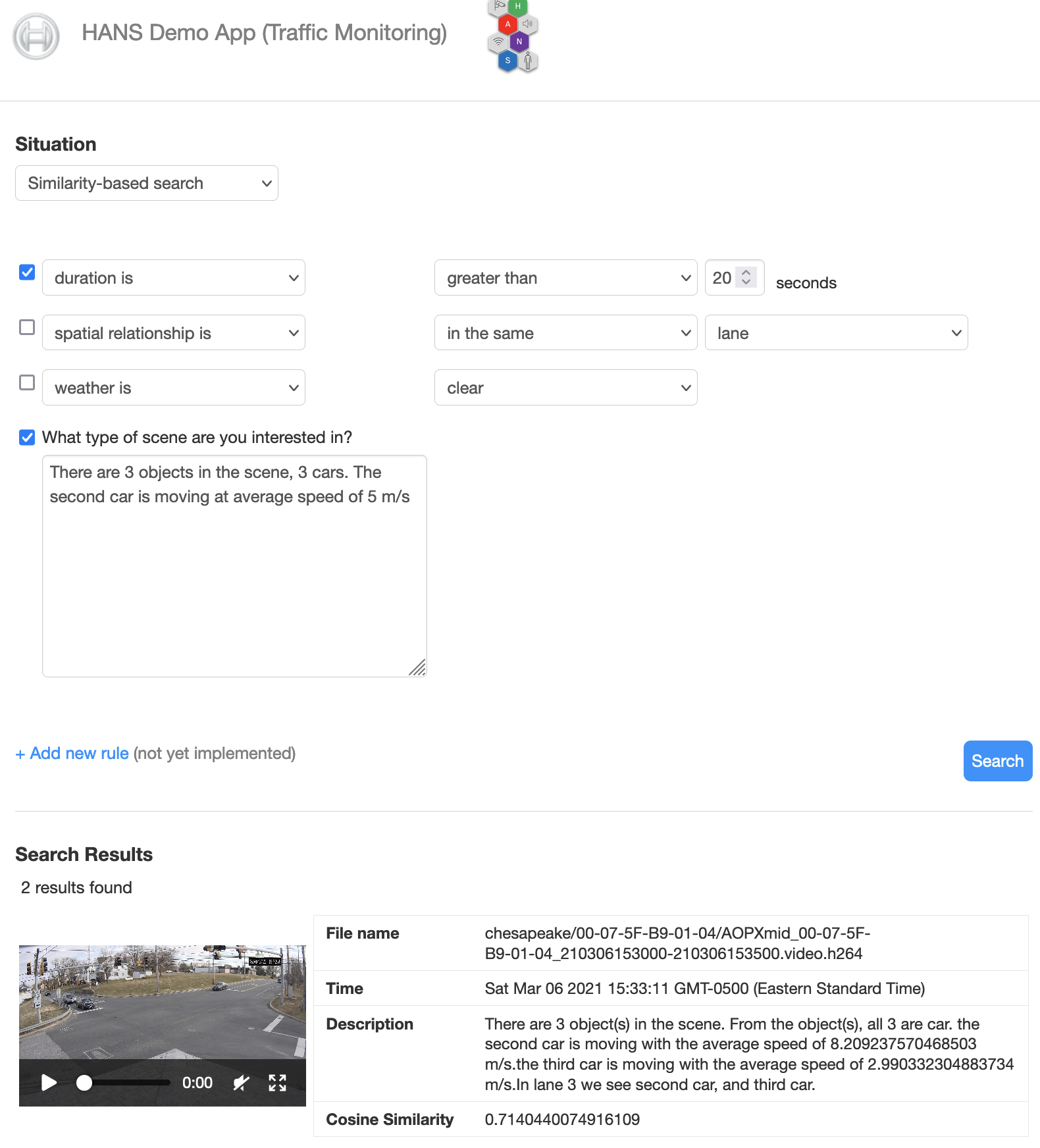}
    \caption{\Ha GUI for intelligent traffic monitoring}
    \label{fig:hans-d}
\end{figure*}

With little effort one can extent such symbolic method to include more flexible rules to cover wider range of traffic phenomenon.
A more flexible rule-based method can facilitate the incorporation of additional extracted information from the image.
Figure~\ref{fig:hans-d}, represents and example of this method, implemented using \Ha GUI\@.
This functionality can help the experts to perform on-the-fly annotations to assess the quality/complexity of rules and provides
various options to implement advanced queries.

\paragraph{Scene Query Based on Graph Structure.}
SOTA reasoners that use whole graph structures to learn appropriate features from data can be used to obviate the scalability issues that symbolic methods generally suffer from \cite{lin2019kagnet,gamanayake2020cluster,scarselli2008graph,gori2005new}.
To this end, we used the \TMKG graph structure of each video frame to find similar frames using methods such as \cite{weinberger2009feature} or \cite{gamanayake2020cluster}.
Table~\ref{tab:hashing-results} presents two examples from the graph-based reasoning model that uses \cite{weinberger2009feature} on top of
\Ha for the scene similarity search.
Here, the query is a node in \Ha associated with a video frame and the output are the frames from other cameras that are closely matched with the scene.
In the current system demonstrator, we rely on the textual description of the frame for both query and output frames.

\begin{table}[ht]
     \small
     \centering
     \begin{tabular}{p{0.43\linewidth} | p{0.43\linewidth}}
         \toprule
         \textbf{Query Description} & \textbf{Best Match Description}\\
         \midrule
         3 objects in the scene, 2 cars and 1 bike.
         The cars are located in lane 3, and move at average speed of 5 m/s.
         &
         3 object(s) in the scene.
         From the object(s), 1 is a bike, and 2 are car.
         the first car is moving with the average speed of 7.74 m/s.
         The second car is moving with the average speed of 5.31 m/s.
         In lane 3 we see first car, and second car.
         \\
         3 objects in the scene, 2 cars and 1 bike.
         The second car is moving at average speed of 5 m/s, in lane 3
         &
         3 object(s) in the scene.
         From the object(s), 1 is a bike, and 2 are car.
         the second car is moving with the average speed of 5.52 m/s.
         In lane 3 we see second car.
         \\
         \bottomrule
     \end{tabular}
     \caption{Scene similarity queries on top of \Ha using feature hashing \protect\cite{weinberger2009feature}.}
     \label{tab:hashing-results}
\end{table}

\paragraph{Natural Language Scene Query.}
Graph based reasoning methods are limited to graph structure and incapable of easily incorporating background knowledge of traffic domain. 
Alternatively, another class of SOTA \textit{reasoners} that \Ha can easily integrate with, relies on language models~\cite{ma2019towards-hykas,liu2019roberta}.
For integration with these, we have implemented conversion methods to transform \TMKG and \CSKG triples into natural language statements.
The resulting sentences include basic visual information such as scene composition, e.g.~\QT{it has 3 cars and 2 trucks}, topological information, e.g.~\QT{road has three lanes} for scene information, or background knowledge, e.g. \QT{truck is a vehicle different from a car}.
One then can utilize these human-readable sentences as input to language models to reason over and \textit{interpret} text, such as questions.
For our the \ITMo use case, we fine-tuned RoBERTa~\cite{liu2019roberta} language model on the background and commonsense statements and used the embedding it provides to find frames that match the embedding of a natural language query.
Accordingly, the examples in Table~\ref{tab:roberta-results} show that the model can \textit{understand} the features associated with different types of entities, applying constraints on their intrinsic properties (e.g., speed) and extrinsic properties (e.g., lane-based localization).\footnote{Note that the order of the entities (i.e., \textit{first}, \textit{second}, \textit{third}) is based on the time stamp associated with perceptual information.}
Here, the first example illustrates natural interaction on exclusion of objects, the second example illustrates a natural interaction based on quality of the objects.

\begin{table}[ht]
     \small
     \centering
     \begin{tabular}{p{0.35\linewidth} | p{0.55\linewidth}}
         \toprule
         \textbf{Query} & \textbf{Best Match Description}\\
         \midrule
         show me a scene with no cars in it
         & There are 2 objects in the scene.
         From the object(s), all 2 are bike.
         The first bike is moving with the average speed of … \\
         show me a scene with a fast moving car
         & There is 1 object in the scene.
         From the object(s), 1 is a car.
         The first car is moving with the average speed of … \\
         \bottomrule
     \end{tabular}
     \caption{Natural language queries on top of \Ha using RoBERTa language model fine-tuned on traffic background knowledge}
     \label{tab:roberta-results}
\end{table}


%% file: Sections-demo/Conclusions.tex
In this paper, we formulate the problem of \ITMo, and introduced \Ha, the Human-Assisted Neuro-Symbolic Architecture for Multi-modal Context Understanding, focusing on its application to intelligent traffic monitoring.
At its core, our solution is centered on a knowledge graph that integrates the data generated by computer vision algorithms (IVA) with general, publicly available knowledge: this method plays a key role in satisfying the requirements 2-5 described in section \ref{sec:problem}. 
We showcase \Ha by integrating various types of knowledge and reasoning architectures for \ITMo tasks such as congestion detection.
Future work on \Ha involves fixing current limitations of the architecture and extensive evaluation.  We are in the process of extending the system with data from multiple cameras and additional sensors. This will allow us to more adequately address multi-modal information fusion and robustness.